\documentclass[10pt]{article}

\usepackage[letterpaper,margin=1in]{geometry}  
\usepackage{setspace}
\usepackage[T1]{fontenc}
\usepackage{times}        
\usepackage{mathptmx}     
\usepackage{bm}           
\usepackage{textcomp}     
\usepackage{pifont}
\usepackage{graphicx}
\usepackage{subcaption}
\usepackage{threeparttable}
\usepackage{makecell}

\usepackage{amsmath,amssymb,amsfonts}
\usepackage{algorithm,algorithmic}

\usepackage{cite}

\usepackage{capt-of}

\newcommand{\BibTeX}{B\kern-.05em{\sc i\kern-.025em b}\kern-.08em T\kern-.1667em\lower.7ex\hbox{E}\kern-.125emX}

\begin{document}

\title{FlexICL: A Flexible Visual In-context Learning Framework for Elbow and Wrist Ultrasound Segmentation}

\author{Yuyue Zhou$^1$, Jessica Knight$^1$, Shrimanti Ghosh$^1$, Banafshe Felfeliyan$^1$, \\ Jacob L. Jaremko$^1$, Abhilash R. Hareendranathan$^1$ \\
$^1$Department of Radiology and Diagnostic Imaging, University of Alberta, \\ Edmonton, AB, Canada
}

\date{}

\maketitle

\begin{abstract}
Elbow and wrist fractures are the most common fractures in pediatric populations. Automatic segmentation of musculoskeletal structures in ultrasound (US) can improve diagnostic accuracy and treatment planning. Fractures appear as cortical defects but require expert interpretation. Deep learning (DL) can provide real-time feedback and highlight key structures, helping lightly trained users perform exams more confidently. However, pixel-wise expert annotations for training remain time-consuming and costly.
To address this challenge, we propose FlexICL, a novel and flexible in-context learning (ICL) framework for segmenting bony regions in US images. We apply it to an intra-video segmentation setting, where experts annotate only a small subset of frames, and the model segments unseen frames. We systematically investigate various image concatenation techniques and training strategies for visual ICL and introduce novel concatenation methods that significantly enhance model performance with limited labeled data. By integrating multiple augmentation strategies, FlexICL achieves robust segmentation performance across four wrist and elbow US datasets while requiring only 5\% of the training images. It outperforms state-of-the-art visual ICL models like Painter, MAE-VQGAN, and conventional segmentation models like U-Net and TransUNet by 1-27\% Dice coefficient on 1,252 US sweeps. These initial results highlight the potential of FlexICL as an efficient and scalable solution for US image segmentation well suited for medical imaging use cases where labeled data is scarce.

\end{abstract}


\section{Introduction}
Pediatric emergency departments (EDs)  in Canada are increasingly burdened with high patient volumes and limited resources, resulting in average wait times exceeding four hours for children presenting with distal radius (wrist) and supracondylar (elbow) injuries \cite{xiemichaelAugmentingLowCostPoint2025}. Accurate diagnosis typically requires a sequence of clinical examinations and imaging studies to determine the presence of fractures. However, clinical data suggest that approximately 45–85\% of pediatric patients presenting with wrist or elbow injuries are ultimately found not to have fractures \cite{knight2D3DUltrasound2023d, knight2DVs3D2023a}. This results in many children undergoing prolonged and often distressing wait times for conditions that may not require urgent intervention. These challenges underscore the need for more efficient, automated diagnostic tools to assist in triaging and evaluating musculoskeletal injuries in pediatric emergency settings.

\subsection{Elbow and wrist fracture}
Elbow and wrist fractures are among the most common pediatric fractures, accounting for approximately 15\% and 33\% of all fractures in children, respectively \cite{brudvikChildhoodFracturesBergen2003b, hedstromEpidemiologyFracturesChildren2010c, chengLimbFracturePattern1993b, saeedElbowFracturesOverview2025}. These injuries typically result from accidental falls and are frequently associated with pain, swelling, tenderness, and restricted joint mobility \cite{saeedElbowFracturesOverview2025}. Given the variability in fracture severity, location, and complexity, timely and accurate diagnosis is critical for optimizing treatment outcomes \cite{skaggsElbowFracturesChildren1997a}. While radiography remains the gold standard for fracture identification, its reliance on specialized equipment and radiation exposure restrict its availability to hospitals and well-equipped medical facilities.

Point-of-care ultrasound (POCUS) is a fast, radiation-free, portable, and cost-effective alternative to X-rays, suitable for EDs, clinics, ambulances, and remote locations. Lightly trained users, such as emergency physicians and triage nurses, can perform scans, with studies showing diagnostic accuracy comparable to or exceeding radiographs for wrist and elbow fractures \cite{knight2D3DUltrasound2023d, knight2DVs3D2023a}. 

\begin{figure*}[t]
    \centering
    \begin{subfigure}[b]{0.32\textwidth}
        \centering
        \includegraphics[width=\linewidth]{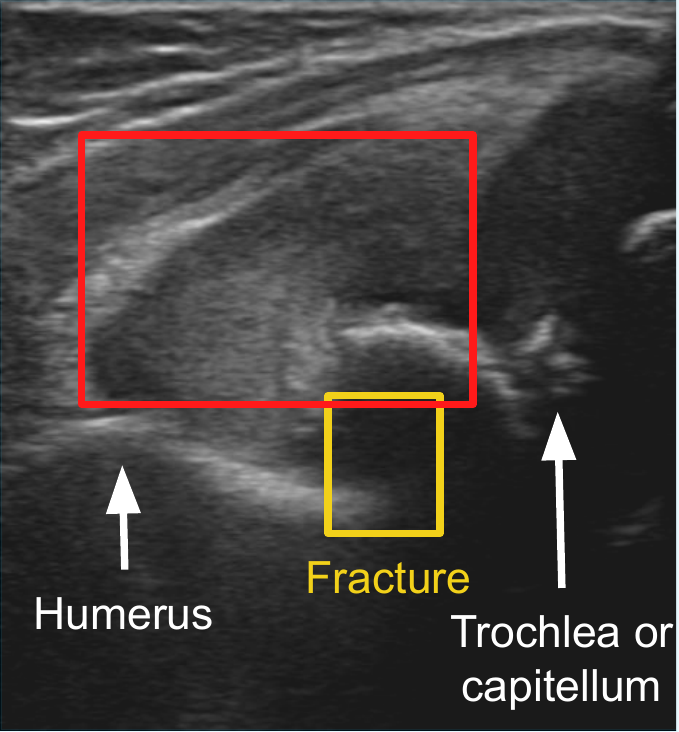}
        \caption{}
        \label{fig1:a}
    \end{subfigure}
    \hspace{3mm}
    \begin{subfigure}[b]{0.32\textwidth}
        \centering
        \includegraphics[width=\linewidth]{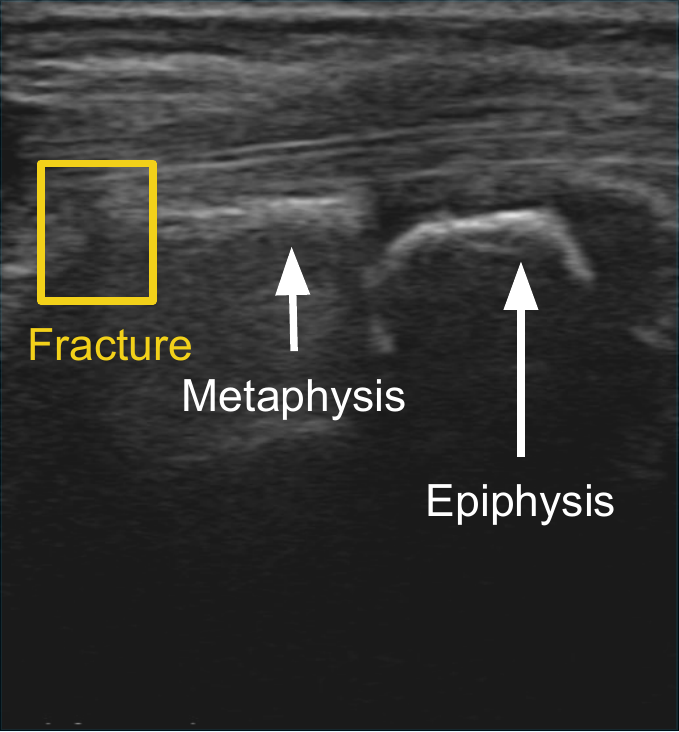}
        \caption{}
        \label{fig1:b}
    \end{subfigure}
    \caption{\textbf{(a) Elbow US image showing humerus, region with effusion (inside the red box), a fracture (inside the yellow box), trochlea or capitellum, and ulna or olecranon. (b) Wrist US image showing bony regions like metaphysis, epiphysis, and a fracture (inside the yellow box).}}
    \label{fig1}
\end{figure*}

Despite its advantages, ultrasound (US) adoption for fracture detection is limited by noisy images, blurred edges, and artifacts that make interpretation difficult for non-experts (Fig. ~\ref{fig1}). Improving image clarity and diagnostic reliability is crucial for broader use in point-of-care and resource-limited settings.

\subsection{Artificial Intelligence Segmentation}
Advances in artificial intelligence (AI) offer new opportunities to automate medical image interpretation, including US, which typically requires expert knowledge. Automatic AI segmentation of clinically relevant structures is a crucial first step in diagnosis \cite{pinto-coelhoHowArtificialIntelligence2023}. Traditional medical image segmentation relies heavily on manual annotation, which is time-consuming as each video can contain hundreds of frames and patients often need multiple scans from different angles. The difficulty is compounded by the fact that non-experts cannot interpret anatomical structures in US images, increasing the cost and complexity of labeling. Besides, the large domain gap between natural and medical US images limits the use of pre-trained models like SAM \cite{kirillovSegmentAnything2023a}, making reliance on custom models and expert annotations a major barrier to scalable AI adoption.

\begin{figure*}[t]
    \centering
    \includegraphics[width=\linewidth]{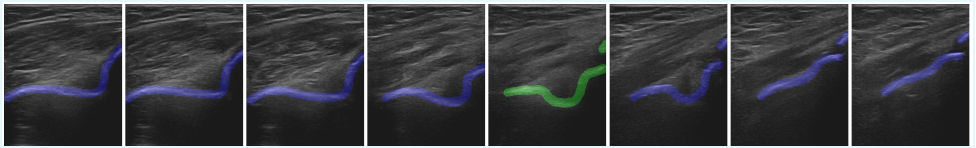}
    \caption{\textbf{The proposed low-annotation intra-video segmentation framework for musculoskeletal US. Sparse manual annotations are required (shown in green), and the model automatically generates segmentations for the remaining frames (shown in blue).}}
    \label{fig2}
\end{figure*}

To address these challenges, we propose a visual in-context learning (ICL) framework for intra-video ultrasound segmentation of elbow and wrist bony regions. Our approach learns from a small subset of annotated frames to segment the remaining frames automatically (Fig.~\ref{fig2}), reducing reliance on extensive manual labeling while maintaining clinically relevant accuracy.

\subsection{Contributions}
The main contributions and novelty of this work are:

(1) We propose a novel and streamlined \textbf{flexible} visual \textbf{ICL} framework named FlexICL, substantially simpler than existing state-of-the-art models, consisting of a ViT-Base encoder followed by a lightweight 1-layer CNN decoder.

(2) We are the first to systematically explore various image augmentation strategies of image concatenation for visual ICL.
Our findings highlight techniques that significantly enhance segmentation performance by encouraging the model to focus on contextual information rather than merely memorizing the data.

(3) We conduct extensive comparisons with both state-of-the-art visual ICL models and conventional segmentation methods across four US datasets (Elbow 2D/3D and Wrist 2D/3D). The results demonstrate consistently superior performance from our model.

\section{ Related work}
\subsection{Precise image segmentation with limited labels}
Image segmentation requires the highest level of supervision, making it highly sensitive to annotation quality and quantity \cite{tajbakhshEmbracingImperfectDatasets2020}. Obtaining pixel-level labels is labor-intensive and costly, especially in medical imaging where expert annotation is needed and datasets are limited \cite{guanDomainAdaptationMedical2022}. This highlights the need for models that perform well with smaller labeled datasets. To address limited data, researchers have explored strategies from simple data augmentation and regularization to more advanced deep learning methods.

Transfer learning, one of the most widely used approaches, enables models to apply knowledge from large pre-trained datasets to related tasks, allowing faster and more efficient training on smaller datasets. However, it still presents challenges. Key limitations include domain mismatch and overfitting on small datasets \cite{alzubaidiNovelTransferLearning2021}. Domain shifts are especially pronounced in medical imaging, where feature distributions vary across modalities. Studies \cite{zoetmulderDomainTaskspecificTransfer2022,albadawyDeepLearningSegmentation2018,poochCanWeTrust2020} show that domain gaps can significantly degrade model performance, with the severity linked to dataset distribution differences. Addressing these gaps is essential for effective adaptation of pre-trained models in medical imaging.

Semi-supervised learning leverages both labeled and unlabeled data using methods like pseudo-labeling and consistency regularization. Its performance, however, is highly dependent on pseudo-label quality and dataset balance \cite{kimDistributionAligningRefinery2021,wuExploringFeatureRepresentation2024}. To improve model performance, some studies have proposed methods for refining pseudo-labels, including the use of similarity matrices \cite{liuSemiSupervisedMedicalImage2024}, exploring pseudo-label thresholds \cite{zengSSTBNSemiSupervisedTriBranch2023}, and evaluating confidence levels \cite{yaoEnhancingPseudoLabel2022}. Despite these advancements, the additional computational burden remains a significant challenge.

Few-shot learning enables models to adapt with limited labeled data using approaches like meta-learning and transfer learning. However, it often suffers from generalization and overfitting issues, requiring specialized architectures or regularization that increase computational cost \cite{nieCrossDomainFewShotSegmentation2024}.

Self-supervised learning (SSL) leverages large unlabeled datasets through a two-stage process: pretraining on a pretext task to learn latent features, followed by fine-tuning on limited labeled data \cite{balestrieroCookbookSelfSupervisedLearning2023}. Recent studies have demonstrated the efficacy of SSL in medical imaging segmentation \cite{felfeliyanSelfsupervisedRCNNMedicalImage2023b, almalkiSelfSupervisedLearningMasked2023a}, where labeled datasets are often scarce. However, the success of SSL heavily depends on the careful selection of an appropriate pretext task \cite{zaiemPretextTasksSelection2021, albelwiSurveySelfSupervisedLearning2022}. Furthermore, the two-stage training process poses challenges, as it requires significant computational resources and extended training time \cite{zhouSelfSupervisedLearningMore2023c}.

Large models like SAM \cite{kirillovSegmentAnything2023a}, SAM 2 \cite{raviSAMSegmentAnything2024} perform well across domains with minimal task-specific data and training. However, they tend to be less effective for domain-specific tasks, such as medical imaging, where specialized knowledge is critical \cite{heComputerVisionBenchmarkSegmentAnything2023a, huangSegmentAnythingModel2024}. 

Finally, several researchers have applied fully supervised models, such as U-Net \cite{ronnebergerUNetConvolutionalNetworks2015d}, to segmentation tasks with limited labeled data. Their study observed that model performance improved noticeably with increasing data size but plateaued beyond a certain point \cite{bardisDeepLearningLimited2020}.

\subsection{Visual in-context learning}
In-context learning (ICL) is an emerging NLP paradigm that enables models to adapt to tasks using contextual information. Inspired by human learning, it has recently been extended to the visual domain, known as Visual ICL, which generally follows two main strategies.

(1) Adding learnable visual prompts to images while keeping models frozen. This method adds a border of learnable visual prompts to input images while keeping the vision model frozen. The prompts provide task-specific context, allowing adaptation to new tasks without changing the model’s core parameters. This approach has been explored by Zhang et al. \cite{zhangInstructMeMore2024b}, Bahng et al. \cite{bahngExploringVisualPrompts2022b}, Huang et al. \cite{huangPromptingUnseenDetecting2024}, Ren et al. \cite{renAreYouCopying2024}, and Wu et al. \cite{wuUnleashingPowerVisual2023a}, demonstrating its effectiveness in guiding frozen models to perform vision tasks.

(2) Utilization of the support-query pair. This method utilizes a support-query pair structure to guide the model’s predictions. Specifically, the model is presented with a "support pair," which includes an input image and its corresponding output, as well as a "query pair," where the output image is missing. Specifically, this method can be divided into two branches: 

(2.1)The model, similar to few-shot learning, encodes support pairs and query inputs separately before fusing them to generate outputs. Leveraging visual context from support pairs, it enables rapid adaptation to new tasks and has shown improved performance across image segmentation \cite{hossainVisualPromptingGeneralized2024}, image editing \cite{zhaoInstructBrushLearningAttentionbased2024}, image classification \cite{vettoruzzoUnsupervisedMetaLearningContext2025}, image restoration \cite{rajagopalanAWRaCLeAllWeatherImage2024} and beyond.

(2.2) Framing computer vision tasks as image inpainting problems.  First introduced by Amir Bar et al. \cite{barVisualPromptingImage}, this method utilizes a concatenated support-query pair image to guide the model’s predictions. Specifically, the input image is the concatenation of four images: support input, support output, query input, and query output (often marked as back region for missing, Fig. ~\ref{fig5}). The model’s task is to inpaint the blacked-out region, effectively generating the query output by learning from the context provided by the support pair. The MAE-VQGAN model combines masked autoencoders (MAE) with VQGAN to improve inpainting, inspiring further research in masked image modeling (MIM) and visual ICL. By using inpainting to incorporate contextual information, Visual ICL enhances task generalization across vision challenges. Wang et al. \cite{wangImagesSpeakImages2023b} proposed Painter, leveraging MIM with Vision Transformers (ViT) \cite{dosovitskiyImageWorth16x162021c} to rapidly adapt to tasks including segmentation, depth estimation, and keypoint detection. Sun et al. \cite{sunExploringEffectiveFactors2023b} focused on utilizing the large scale vision model for inpainting-based segmentation. Wang et al. \cite{wangSegGPTSegmentingEverything2023b} introduced SegGPT, a MIM-ViT model that segments diverse objects using in-context examples, demonstrating adaptability across segmentation tasks. Liu et al. \cite{liuUnifyingImageProcessing2024a} proposed PromptGIP, a universal framework combining MIM and Transformer for general image processing tasks such as restoration, enhancement, and feature extraction, enabling cross-domain task adaptation without task-specific fine-tuning. Zhang et al. \cite{zhangWhatMakesGood2023b} and Xu el al. \cite{xuGlobalOptimalVisual2024} proposed methods for retrieving the most similar support pair for a given query pair in visual ICL models, significantly enhancing the model's performance.

Despite extensive research on natural images, visual ICL via image inpainting in medical imaging is limited. Kumar et al. \cite{kumarVisualContextLearning2024} showed SegGPT’s potential for eczema segmentation using nearest-neighbor support-query retrieval. Similarly, Wu et al. \cite{wuTumorSegmentationWhole2024} applied SegGPT to whole slide image segmentation, demonstrating that carefully selected support images can outperform full fine-tuning. Additionally, Zhou et al. \cite{zhouSimICLSimpleVisual2024} introduced a visual ICL model for segmenting bony regions in wrist US images. These early works demonstrate the potential of visual ICL in advancing medical image segmentation tasks.

\subsection{Our work}
Building upon the concept of image inpainting proposed by Bar et al. \cite{barVisualPromptingImage}, and followed by the work by Zhou et al. \cite{zhouSimICLSimpleVisual2024}, this study presents FlexICL, a novel flexible segmentation framework combining visual ICL with MIM. Using augmented support-query pairs, FlexICL enables flexible segmentation with limited labeled data. We conducted a comprehensive exploration of how the training can be optimized to enhance model performance. We evaluate FlexICL on prospectively collected elbow and wrist US datasets, addressing the challenges posed by limited labeled data, aiming to improve segmentation accuracy.

Unlike prior MAE-VQGAN methods, FlexICL is implemented within the SSL SimMIM framework \cite{xieSimMIMSimpleFramework2022b}, applying random patch masking with ViT and reconstructing images via Mean Absolute Error loss. This is the first study to apply visual ICL to elbow US imaging. Experiments show FlexICL outperforms state-of-the-art segmentation and visual ICL models, eliminating the conventional two-stage pretraining-finetuning process while maintaining high accuracy, highlighting the efficiency and generalizability of MIM-based approaches.

\section{Methods}
\subsection{Datasets collection}
Data was collected prospectively at Stollery Children’s Hospital ED with institutional ethics approval and informed parental consent. We used a Philips Lumify L5-12 MHz pocket-sized POCUS scanner and a Philips iU22 machine with a 13 MHz VL13-5 transducer to collect US videos. US examinations were performed on children aged 0-17 who presented to the ED with elbow or wrist trauma. Our elbow protocol included two US cine sweeps of the elbow in the following locations: (1) the dorsal view over the olecranon fossa, and (2) the volar view over the coronoid and radial fossa. Due to logistical constraints, a few patients did not undergo complete scans, and the details are shown in Table 1. Our wrist protocol included five US cine sweeps in the following locations: (1) dorsal, including metaphysis, epiphysis, and first row of carpal bones; (2) proximal dorsal, including metaphysis and epiphysis; (3) radial, including metaphysis, epiphysis, and first row of carpal bones; (4) volar, including metaphysis, epiphysis, and first row of carpal bones; and (5) proximal volar, including metaphysis and epiphysis. Bony region segmentation masks for each scan were generated by a musculoskeletal sonographer with 10 years of experience supervised by an experienced radiologist, using freeware ITK-Snap. We converted the video into a sequence of single frames, and only frames with distinct and clear views that the sonographer felt to be confident of pathology were labeled and used in our study. Data was randomized and anonymized after collection. The details of datasets can be found in Table~\ref{tab1}.

\begin{table*}[t]
\centering
\caption{Datasets details}
\begin{tabular}{|p{2cm}|p{3cm}|p{3cm}|p{3cm}|p{3cm}|}
\hline
 & Elbow & Elbow & Wrist & Wrist \\
\hline
Video type & 2D US & 3D US & 2D US & 3D US \\
\hline
Collected by
&Philips Lumify L5-12 MHz transducer 
&Philips iU22 machine with a 13 MHz VL13-5 transducer
&Philips Lumify L5-12 MHz transducer 
&Philips iU22 machine with a 13 MHz VL13-5 transducer\\
\hline
Total \# of Patients
&53
&50
&111
&104\\
\hline
\# of Sweep videos
&101
&95
&543
&513\\
\hline
\# of Images 
&9731
&8599
&24902
&23703\\
\hline
\# of Patients without complete scans
&6
&5
&10
&4
\\
\hline
\end{tabular}
\label{tab1}
\end{table*}

\subsection{Dataset splitting strategy}
To ensure fair and consistent evaluation across different segmentation paradigms, we designed a unified intra-video data splitting strategy. The main goals were (1) to avoid data leakage between training and test sets, (2) to maintain sufficient frame diversity for model generalization, and (3) to ensure comparability between visual ICL and conventional supervised models.

\subsubsection{Overall procedure}
For each video, a small proportion of frames (x\%, e.g., 5\% or 10\%) were randomly selected as the training set, while the remaining (100 – x\%) formed the test set. This intra-video sampling ensures that all models are trained and evaluated on the same underlying video content but with non-overlapping frames.

\subsubsection{Visual ICL setting}
Within the training set, frames were further divided equally into support and query subsets, reflecting the two-part structure of visual prompting. During testing, each query (test) frame was paired with a support frame from the training set based on the closest frame number. This choice follows prior findings that visual prompting performs best when support–query pairs are temporally or visually related. Details can be found in Fig.~\ref{fig3}\subref{fig3:a}.

Furthermore, as the visual ICL models employed in this study (MAE-VQGAN, Painter, and the proposed FlexICL) are built upon the MIM framework, they inherently do not require a separate validation set. These models are pretrained in a self-supervised or prompt-based manner, where learning relies on reconstructive objectives rather than validation-driven optimization. Consequently, model performance is directly evaluated on the test set to assess generalization, consistent with standard practices in MIM-based visual learning.

\subsubsection{Conventional supervised setting}
\begin{figure*}[t]
    \centering
    \begin{subfigure}[b]{0.7\textwidth}
        \centering
        \includegraphics[width=\linewidth]{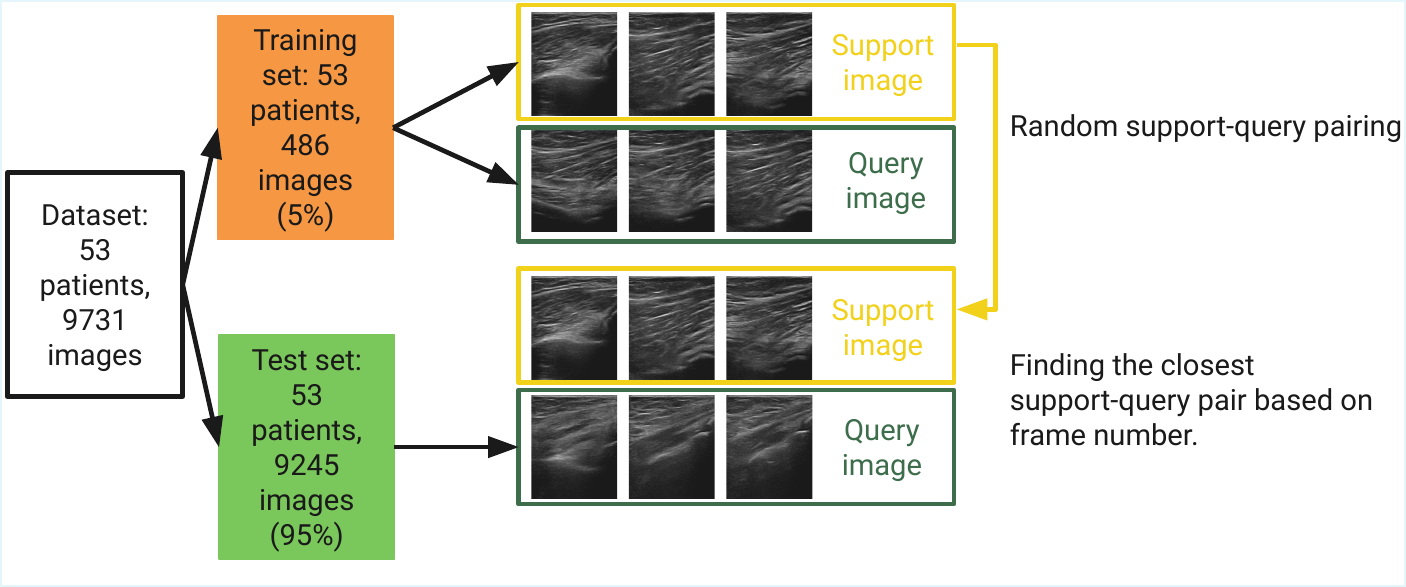}
        \caption{}
        \label{fig3:a}
    \end{subfigure}
    \begin{subfigure}[b]{0.7\textwidth}
        \centering
        \includegraphics[width=\linewidth]{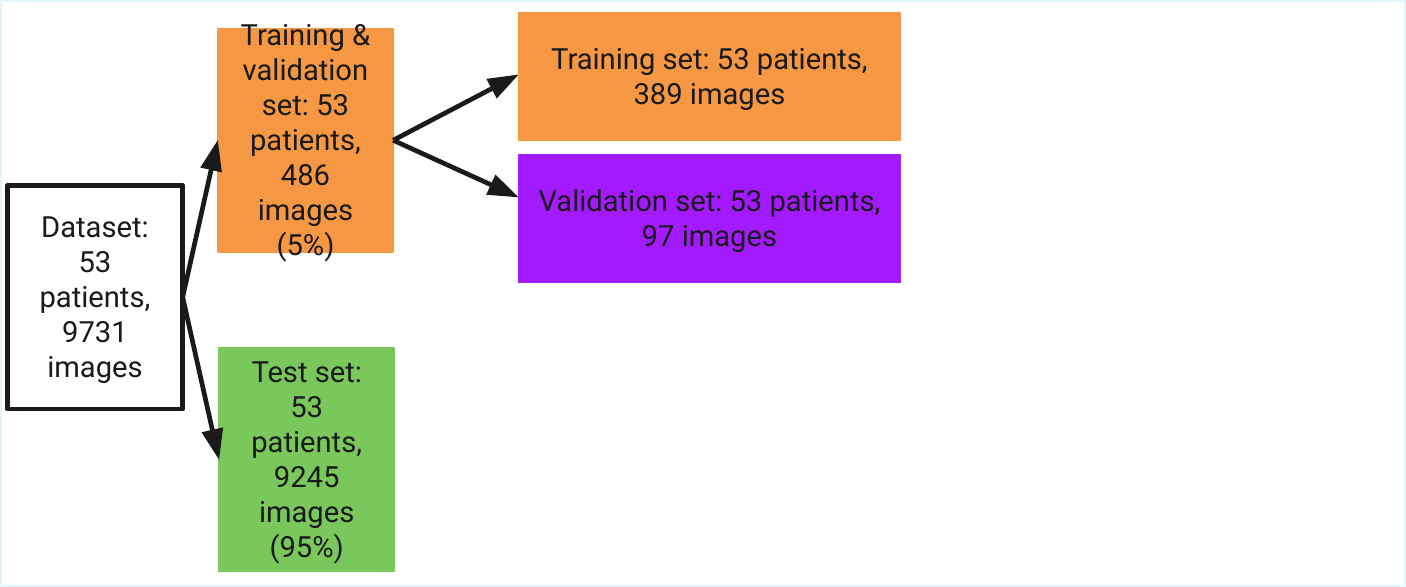}
        \caption{}
        \label{fig3:b}
    \end{subfigure}
    \caption{\textbf{Dataset splitting and preprocessing. (a) Visual ICL model. 5\% of frames were randomly selected from each video as the training set and divided into support and query pools. Each support image–mask pair was randomly matched with a query image–mask pair for training. (b) Conventional segmentation model. Training and validation sets were constructed using the same sampling strategy as the Visual ICL model—randomly selecting 5\% of frames from each video. A fixed random seed was used to ensure consistent sampling of training (and validation) sets across different models. Note that there is no image overlap between the training, validation and test sets for both visual ICL and conventional fully supervised models data splitting.}}
    \label{fig3}
\end{figure*}

For supervised segmentation models, the original training set was further split into 80\% for training and 20\% for validation. The model showing the best validation performance was subsequently evaluated on the held-out test set. This 80/20 protocol aligns with standard practice in supervised segmentation tasks and allows for effective hyperparameter tuning while preserving test independence. Details can be found in Fig.~\ref{fig3}\subref{fig3:b}.

This unified yet model-specific splitting strategy balances fairness, comparability, and performance optimization, ensuring coherent evaluation across different segmentation approaches.

\subsection{FlexICL}
\subsubsection{Pairwise Augmentation}
\begin{figure*}[t]
    \centering
    \begin{subfigure}[t]{0.5\textwidth}
        \centering
        \vspace{0pt}
        \includegraphics[width=\linewidth]{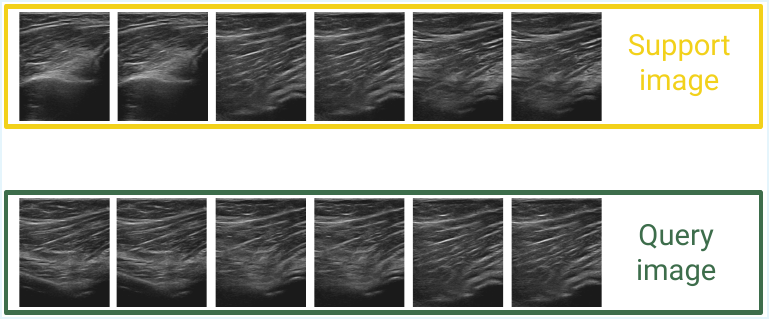}
        \caption{}
        \label{fig4:a}
    \end{subfigure}
    
    \begin{subfigure}[t]{0.4\textwidth}
        \centering
        \vspace{0pt}
        \includegraphics[width=\linewidth]{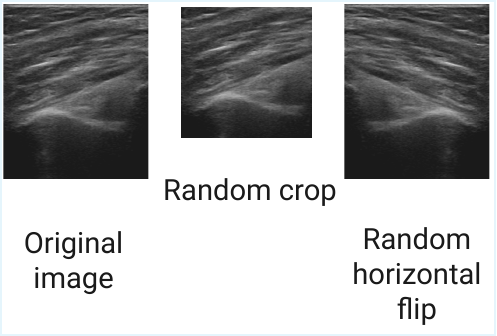}
        \caption{}
        \label{fig4:b}
    \end{subfigure}
    \hspace{2mm}
    \begin{subfigure}[t]{0.4\textwidth}
        \centering
        \vspace{0pt}
        \includegraphics[width=\linewidth]{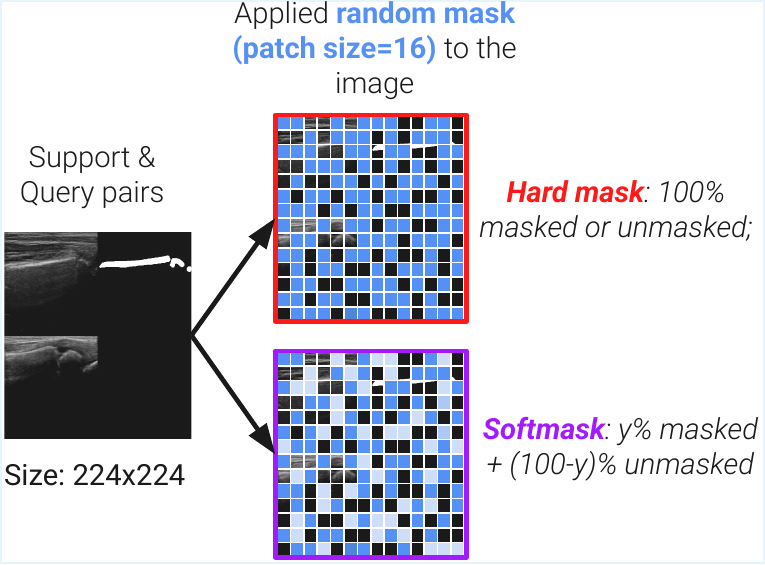}
        \caption{}
        \label{fig4:c}
    \end{subfigure}
    \caption{\textbf{Different training strategies for FlexICL. (a) Pairwise augmentation: Illustrated with a 2× augmentation example, where support and query images are duplicated in their respective pools. (b) Image-wise augmentation: The original image undergoes a random crop or horizontal flip before concatenation. (c) Softmask and hardmask random masking: Demonstrating the application of random masking techniques.}}
    \label{fig4}
\end{figure*}

All images and ground truth segmentation were preprocessed by paddling them to a square format and resizing them to 224 × 224. During training, we augmented the support-query pairs n times (e.g., 5, 10, 20) to enhance data diversity. Given m support images and m query images, we initially formed m support-query pairs. To enrich the training data, we repeated this process n times, each time randomly re-pairing support and query images and applying data augmentations. As a result, the total number of training pairs increased from m to n × m, improving the model’s exposure to diverse combinations without requiring additional labeled data. Further details are illustrated in Fig. ~\ref{fig4}\subref{fig4:a}.

\subsubsection{Imagewise Augmentation}
Before concatenation, we applied imagewise augmentations on images and ground truth segmentation during training as shown in Fig. ~\ref{fig4}\subref{fig4:b}, including random horizontal flipping and cropping to a random size between 150 × 150 and 224 × 224, followed by resizing to 224 × 224.

\subsubsection{Softmask and Hardmask Random Masking}
The paired support and query images-ground truth segmentations were concatenated into a single image for model input. Each support pair (support image, support ground truth segmentation) and query pair (query image, query ground truth segmentation) was concatenated and resized to 224 × 224. We randomly masked regions in the concatenated input and fed both masked and unmasked areas to the model. Based on preliminary experiments, we set the mask patch size to 16 and the training mask ratio to 0.6, following our previous research \cite{zhouSimICLSimpleVisual2024}.

For hard masking, fully masked regions were entirely replaced with trainable parameters, whereas unmasked regions remained unchanged. In contrast, we explored a soft mask strategy using a weighted combination approach. Instead of fully replacing masked areas, we applied a blending function: masked region=y\%×trainable parameters+(100-y)\%×original embedding, where y represents the masking ratio. The illustration is shown in Fig. ~\ref{fig4}\subref{fig4:c}.

\subsubsection{Epochwise Augmentation}
To enhance data variability, we randomly shuffled the support-query set at each training epoch, as shown in Fig. ~\ref{fig5}.

\subsubsection{MIM bases FlexICL}
\begin{figure*}[t]
    \centering
    \includegraphics[width=\linewidth]{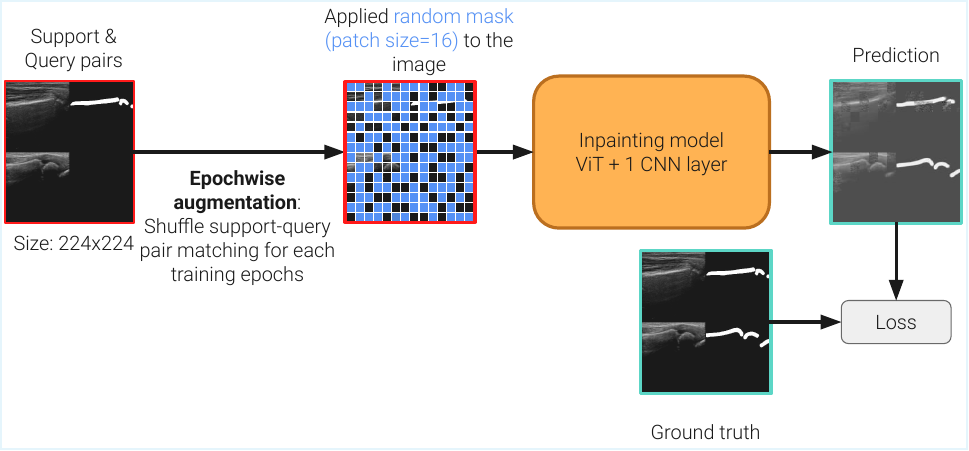}
    \caption{\textbf{FlexICL Model. We applied epoch-wise augmentation, reshuffling support-query pairings in each training epoch.}}
    \label{fig5}
\end{figure*}

Our model follows the SimMIM framework \cite{xieSimMIMSimpleFramework2022b} with a vanilla ViT-Base encoder \cite{dosovitskiyImageWorth16x162021c}, consisting of 12 layers and 768 embedding dimensions. A lightweight prediction head (decoder) with a single convolutional layer was used to reconstruct the masked image while accelerating the training process, in line with SimMIM framework. The model was trained from scratch using Mean Absolute Error loss for reconstruction. Fig. ~\ref{fig5} provides an overview of our approach.

Implementation Details: After hyperparameter tuning, we employed the AdamW optimizer and set the learning rate to 0.0005 and weight decay to 0.05 after hyperparameter tuning. The model was trained for 1,200 epochs with a batch size of 64 on an NVIDIA V100 GPU. Preliminary experiments indicated that the model generally achieved its best or near-optimal performance around this point. 

\subsubsection{Baseline comparison models}
We compared our work with state-of-the-art models: conventional segmentation models U-Net \cite{ronnebergerUNetConvolutionalNetworks2015d}, TransUNet \cite{chenTransUNetTransformersMake2021}, visual ICL models MAE-VQGAN \cite{barVisualPromptingImage} and Painter \cite{wangImagesSpeakImages2023b}. 

For the conventional segmentation models, U-Net (ResNet34 encoder) and TransUNet (ResNet 50 encoder + ViT base backbone), we used the SGD optimizer with a learning rate of 0.005, weight decay of 0.05, and a batch size of 64 after hyperparameter tuning. BCE+Dice loss was used as a loss function. Both models were trained for 200 epochs initialized with ResNet encoder weights pretrained on ImageNet, and models with the smallest loss on validation set were saved as the best model.

For MAE-VQGAN, we adopted the ViT-Large backbone and fine-tuned the model for 600 epochs based on performance. We used the AdamW optimizer, following the original paper. After hyperparameter tuning, we set the learning rate to 0.001, the weight decay to 0.05, and the batch size to 64. Pretrained weights (from ImageNet training for 1000 epochs) were provided by the original authors. Other settings including loss function were kept consistent with the original configuration.

For the Painter model, we also used the AdamW optimizer as in the original implementation, with a learning rate of 0.0001, weight decay of 0.05 after hyperparameter tuning, and a batch size of 64. The model was trained for 1200 epochs from scratch based on best performance. Other training settings including loss function followed the original paper.

\subsection{Evaluation metrics}
Dice similarity coefficient (DSC = 2TP/(2TP + FP+FN)) and Jaccard index (IoU = TP/(TP + FP+FN)) were used as evaluation metrics.   

\section{Results}
\subsection{Effect of Training Data Size on Model Performance}
To investigate the impact of training set size on model performance, we trained our vallina visual ICL model using varying fractions of the elbow 2D US dataset. During training, only epochwise augmentation was applied—excluding pairwise and imagewise augmentations—and the default hard mask setting was used. Support-query pairs were randomly shuffled at each epoch to ensure diversity. The results, summarized in Table~\ref{tab2}, demonstrate that without additional augmentations, the visual ICL model underperforms significantly compared to U-Net when trained on small datasets. However, as the proportion of training data increases, the performance of the visual ICL model improves markedly, particularly when the training set reaches 20\% of the full dataset.

\begin{table*}[t]
\centering
\caption{The effect of the ratio of training images for visual ICL model on elbow 2D dataset}
\begin{tabular}{|p{2cm}|p{2cm}|p{2cm}|p{2cm}|p{2cm}|}
\hline
Ratio of images used for training
(\# of images)
&1\%(97)
&5\%(487)
&20\%(1946)
&U-Net, 5\%(487)\\
\hline
DSC
&0.10
&0.20
&0.83
&0.79\\
\hline
IoU
&0.05
&0.11
&0.72
&0.67\\
\hline
\end{tabular}
\label{tab2}
\end{table*}

\subsection{Data augmentation strategies}
Following the protocol established in previous research \cite{zhouSimICLSimpleVisual2024}, we adopt a 60\% masking strategy for all subsequent experiments. We investigate the effects of three types of data augmentation: pairwise augmentation, imagewise augmentation, and epochwise augmentation. All augmentation experiments were conducted using 5\% of the labeled dataset to simulate a low-data regime.

As shown in columns 1, 3, 6, and 7 of Table ~\ref{tab3}, pairwise augmentation significantly improves model performance. However, the benefit plateaus after a certain point, specifically, applying the augmentation five times yields the most substantial improvement, while further increasing it to ten or twenty times does not result in additional gains and, in fact, slightly degrades performance.

Epoch-wise augmentation does not consistently enhance performance (columns 2-3 in Table ~\ref{tab3}), but can lead to improvements in certain scenarios (columns 4-5 in Table ~\ref{tab3}), suggesting its effect may be context-dependent.

Imagewise augmentation consistently improves performance across multiple settings (columns 2–5 in Table ~\ref{tab3}). To verify that gains are due to visual ICL rather than the ViT-CNN model alone, we compared the same ViT-CNN under identical training without the image concatenation of visual ICL. As shown in columns 10–11 in Table ~\ref{tab3}, the standalone ViT-CNN performs poorly on the 5\% labeled segmentation task, indicating a performance collapse.

To further investigate the role of imagewise augmentation, we compared visual ICL with U-Net and ViT-CNN segmentation models, both with and without imagewise augmentation. On U-Net, imagewise augmentation yields no significant improvement (columns 8-9 in Table ~\ref{tab3}), while on ViT-CNN, it leads to performance collapse (columns 10-11 in Table ~\ref{tab3}). These findings highlight the unique compatibility and effectiveness of imagewise augmentation within the visual ICL framework.
\begin{table*}[t]
\centering
\begin{threeparttable}
\caption{Impact of Different Augmentation Strategies on Visual ICL Performance (5\% Labeled Elbow 2D Data)}
\label{tab3}
\begin{tabular}{|p{2cm}|p{0.9cm}|p{0.9cm}|p{0.9cm}|p{0.9cm}|p{0.9cm}|p{0.9cm}|p{0.9cm}|p{0.9cm}|p{0.9cm}|p{0.9cm}|p{0.9cm}|}
\hline
&1
&2
&3
&4
&5
&6
&7
&8
&9
&10
&11\\
\hline
Pairwise augmentation
&1x
&5x
&5x
&\textbf{5x}
&5x
&10x
&20x
&U-Net
&U-Net
&ViT-CNN\tnote{*}
&ViT-CNN\tnote{*}\\
\hline
Imagewise augmentation
&\ding{55}
&\ding{55}
&\ding{55}
&\ding{51}
&\ding{51}
&\ding{55}
&\ding{55}
&\ding{55}
&\ding{51}
&\ding{55}
&\ding{51}\\
\hline
Epochwise augmentation
&\ding{51}
&\ding{55}
&\ding{51}
&\ding{51}
&\ding{55}
&\ding{51}
&\ding{51}
&NA
&NA
&NA
&NA\\
\hline
Mask
&Hard
&Hard
&Hard
&\textbf{Hard}
&Hard
&Hard
&Hard
&NA
&NA
&NA
&NA\\
\hline
Mask ratio
&0.6
&0.6
&0.6
&\textbf{0.6}
&0.6
&0.6
&0.6
&NA
&NA
&NA
&NA\\
\hline
DSC
&0.20
&0.75
&0.75
&\textbf{0.80}
&0.77
&0.73
&0.71
&0.79
&0.79
&0.13
&0.0\\
\hline
IoU
&0.11
&0.62
&0.62
&\textbf{0.68}
&0.65
&0.6
&0.58
&0.67
&0.67
&0.07
&0.0\\
\hline
\end{tabular}
\begin{tablenotes}
\item[*] The ViT-CNN model adopts the same architecture as FlexICL, comprising a ViT-Base encoder and a single-layer CNN decoder.
\end{tablenotes}
\end{threeparttable}

\end{table*}

\subsection{Ratio of Imagewise augmentation}
We conducted experiments to determine the optimal ratio for imagewise augmentation, that is, the proportion of images subjected to imagewise augmentation prior to support-query concatenation. All experiments were performed using 5\% of the training data, with 5× pairwise augmentation, epochwise augmentation, and hard mask settings applied consistently.

The results, presented in Table~\ref{tab4}, indicate that model performance improves as the augmentation ratio increases, up to a certain point. Performance peaks at a ratio of 0.5, after which further increases lead to a decline in accuracy. This suggests that a 50\% imagewise augmentation ratio yields the best segmentation performance under the given conditions.

\begin{table*}[t]
\centering
\caption{Effect of Imagewise Augmentation Ratios on Segmentation Performance (5\% Labeled Elbow 2D Data)}
\begin{tabular}{|p{3cm}|p{2cm}|p{2cm}|p{2cm}|p{2cm}|}
\hline
Pairwise augmentation
&5x
&5x
&\textbf{5x}
&5x\\
\hline
Imagewise augmentation
&\ding{55}
&\ding{51}
&\ding{51}
&\ding{51}\\
\hline
Epochwise augmentation
&\ding{51}
&\ding{51}
&\ding{51}
&\ding{51}\\
\hline
Mask
&Hard
&Hard
&\textbf{Hard}
&Hard\\
\hline
Mask ratio
&0.6
&0.6
&\textbf{0.6}
&0.6\\
\hline
Imagewise augmentation ratio
&0
&0.25
&\textbf{0.5}
&0.75\\
\hline
DSC
&0.75
&0.77
&\textbf{0.8}
&0.77\\
\hline
IoU
&0.62
&0.65
&\textbf{0.68}
&0.64\\
\hline
\end{tabular}
\label{tab4}
\end{table*}

\subsection{ Imagewise augmentation selection}
We evaluated the effects of different imagewise augmentation strategies on the 5\% training set, while keeping the testing set unchanged (i.e., without any augmentation). The results are presented in Table ~\ref{tab5}.

Interestingly, applying random horizontal flipping alone slightly decreased model performance compared to training without any augmentation. In contrast, applying random cropping led to an improvement in performance. Notably, combining random cropping and flipping resulted in the best segmentation performance, highlighting the complementary benefits of these two augmentation techniques when used together.  

\begin{table*}[t]
\centering
\caption{Impact of Imagewise Augmentation Combinations on Segmentation Accuracy (5\% Labeled Elbow 2D Data)}
\begin{tabular}{|p{3cm}|p{2cm}|p{2cm}|p{2cm}|p{2cm}|}
\hline
Pairwise augmentation
&5x
&5x
&\textbf{5x}
&5x\\
\hline
Imagewise augmentation
&only random flip 
&only random crop
&\textbf{random flip+random crop}
&\ding{55}
\\
\hline
Epochwise augmentation
&\ding{51}
&\ding{51}
&\ding{51}
&\ding{51}\\
\hline
Mask
&Hard
&Hard
&\textbf{Hard}
&Hard\\
\hline
Mask ratio
&0.6
&0.6
&\textbf{0.6}
&0.6\\
\hline
Imagewise augmentation ratio
&0.5
&0.5
&\textbf{0.5}
&0\\
\hline
DSC
&0.72
&0.79
&\textbf{0.80}
&0.75\\
\hline
IoU
&0.58
&0.68
&\textbf{0.68}
&0.62\\
\hline
\end{tabular}
\label{tab5}
\end{table*}

\subsection{Softmask and hard mask selection}

We evaluated the impact of mask supervision type, specifically, hard masks versus soft masks, on model performance using 5\% of the training dataset, with 5× pairwise augmentation applied. The results are presented in Table ~\ref{tab6}.

Training with soft masks leads to a slight decrease in segmentation performance compared to training with hard masks. This suggests that, under low-data conditions, the model benefits more from the sharper and more definitive boundary information provided by hard masks. Soft masks may introduce noise or reduce the model’s ability to learn precise object boundaries in such settings.
\begin{table*}[t]
\centering
\caption{ Comparison of Soft Mask vs. Hard Mask Training Strategies (5\% Labeled Elbow 2D Data)}
\begin{tabular}{|p{3cm}|p{2cm}|p{2cm}|}
\hline
&\textbf{Hard mask}
&Soft mask\\
\hline
Pairwise augmentation
&\textbf{5x}
&5x
\\
\hline
Imagewise augmentation
&\ding{55}
&\ding{55}\\
\hline
Epochwise augmentation
&\ding{51}
&\ding{51}\\
\hline
Mask ratio
&\textbf{0.6}
&0.6\\
\hline
DSC
&\textbf{0.75}
&0.72\\
\hline
IoU
&\textbf{0.62}
&0.58\\
\hline
\end{tabular}
\label{tab6}
\end{table*}

\subsection{Effect of Random Masking During Inference}
We investigated the effect of different mask ratios on model performance during inference time, using hard masks for evaluation on test set. The results, summarized in Table ~\ref{tab7}, are based on experiments conducted with both 5\% and 20\% training set sizes. The best performance was achieved when either no mask or a low mask ratio was applied at test time. This suggests that excessive masking during inference may obscure informative regions and hinder the model’s ability to generalize.

\begin{table*}[ht]
\centering
\caption{Effect of Test-Time Mask Ratios on Segmentation Performance}
\begin{tabular}{|p{1.5cm}|p{1.5cm}|p{1.5cm}|p{1.5cm}|p{1.5cm}|p{1.5cm}|p{1.5cm}|}
\hline
\multicolumn{6}{|c|}{\makecell[l]{5\% training set, with imagewise augmentation, \\ pairwise augmentation and epochwise augmentation, hard mask}} & U-Net, 5\% \\
\hline
Mask ratio
&\textbf{0}
&0.15
&0.3
&0.45
&0.6
&0
\\
\hline
DSC
&\textbf{0.80}
&0.79
&0.77
&0.74
&0.69
&0.74
\\
\hline
IoU
&\textbf{0.68}
&0.67
&0.65
&0.61
&0.56
&0.60
\\
\hline

\multicolumn{6}{|c|}{\makecell[l]{20\% training set, with imagewise augmentation, \\ pairwise augmentation and epochwise augmentation, hard mask}} &  \\

\hline
Mask ratio
&0
&\textbf{0.15}
&0.3
&0.45
&0.6
& \\
\hline 
DSC
&0.80
&\textbf{0.83}
&0.82
&0.79
&0.76
& \\
\hline
IoU
&0.69
&\textbf{0.72}
&0.71
&0.68
&0.64
& \\
\hline
\end{tabular}
\label{tab7}
\end{table*}

\subsection{Comparison with other models on multiple datasets}
From these experiments, we established the final FlexICL training strategy: 0.5-ratio imagewise augmentation, pairwise and epochwise augmentation, and a 0.6 mask ratio with hard masks, with no masking during inference. 

\begin{figure*}[t]
    \centering
    \includegraphics[width=\linewidth]{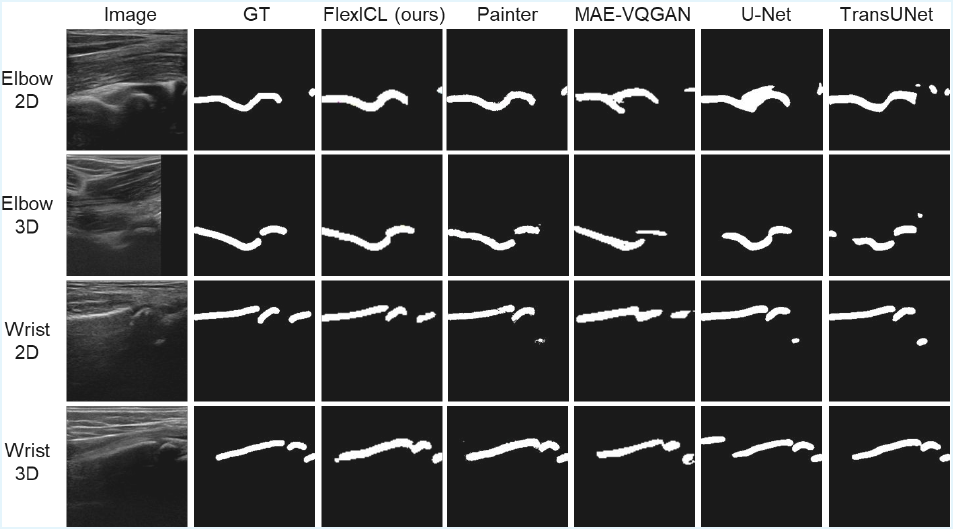}
    \caption{\textbf{Visual comparison of segmentation results across different datasets, using 5\% of the training data.}}
    \label{fig6}
\end{figure*}

To assess generalizability, FlexICL was compared with conventional models (U-Net, TransUNet) and visual ICL models (MAE-VQGAN, Painter) on four US datasets, Elbow 2D/3D and Wrist 2D/3D US, using only 5\% of training data (Table ~\ref{tab8}, Fig. ~\ref{fig6}). 

FlexICL outperformed competing methods in most settings, producing segmentation results visually closer to ground truth, with fewer mis-segmented regions and more accurate bony structure delineation. The Wrist 3D dataset was the only case where it did not achieve top performance, though results remained comparable to the best model. Overall, FlexICL demonstrates robust and superior segmentation in both 2D and 3D US.

\begin{table*}[ht]
\centering
\caption{The comparison with other models on multiple datasets with 5\% training images}
\begin{tabular}{|p{1.5cm}|p{2cm}|p{2cm}|p{2cm}|p{2cm}|p{2cm}|}
\hline
\multicolumn{6}{|l|}{Elbow 2D} \\ 
\hline
Model
&\textbf{FlexICL(ours)}
&Painter \cite{wangImagesSpeakImages2023b}
&MAE-VQGAN \cite{barVisualPromptingImage}
&U-Net \cite{ronnebergerUNetConvolutionalNetworks2015d}
&TransUNet \cite{chenTransUNetTransformersMake2021}\\ 
\hline
DSC
&\textbf{0.80}
&0.68
&0.61
&0.79
&0.73\\ 
\hline
IoU
&\textbf{0.68}
&0.55
&0.47
&0.67
&0.59\\ 
\hline
\multicolumn{6}{|l|}{Elbow 3D}\\ 
\hline
Model
&\textbf{FlexICL(ours)}
&Painter \cite{wangImagesSpeakImages2023b}
&MAE-VQGAN \cite{barVisualPromptingImage}
&U-Net \cite{ronnebergerUNetConvolutionalNetworks2015d}
&TransUNet \cite{chenTransUNetTransformersMake2021}\\ 
\hline
DSC
&\textbf{0.84}
&0.72
&0.64
&0.76
&0.77\\ 
\hline
IoU
&\textbf{0.73}
&0.60
&0.49
&0.63
&0.64\\ 
\hline
\multicolumn{6}{|l|}{Wrist 2D}\\ 
\hline
Model
&\textbf{FlexICL(ours)}
&Painter \cite{wangImagesSpeakImages2023b}
&MAE-VQGAN \cite{barVisualPromptingImage}
&U-Net \cite{ronnebergerUNetConvolutionalNetworks2015d}
&TransUNet \cite{chenTransUNetTransformersMake2021}\\ 
\hline
DSC
&\textbf{0.86}
&0.85
&0.60
&0.84
&0.85\\ 
\hline
IoU
&\textbf{0.77}
&0.75
&0.45
&0.73
&0.75\\ 
\hline
\multicolumn{6}{|l|}{Wrist 3D}\\ 
\hline
Model
&FlexICL(ours)
&\textbf{Painter} \cite{wangImagesSpeakImages2023b}
&MAE-VQGAN \cite{barVisualPromptingImage}
&U-Net \cite{ronnebergerUNetConvolutionalNetworks2015d}
&TransUNet \cite{chenTransUNetTransformersMake2021}\\ 
\hline
DSC
&0.86
&\textbf{0.87}
&0.59
&0.87
&0.86\\
\hline
IoU
&0.77
&\textbf{0.78}
&0.44
&0.77
&0.76
\\
\hline
\end{tabular}
\label{tab8}
\end{table*}

\section{Discussion}
The results show that limited training data hinders the baseline visual ICL model (SimMIM-ViT), as ViT-like other Transformer models, is less suited for segmentation and requires more data than CNN-based models like U-Net (Table ~\ref{tab3}, column 10).

Image concatenation-based augmentation enhances data diversity and improves performance. Pairwise and imagewise augmentations increase training images and visual variety, yielding substantial gainsand ensuring that the model focuses on contextual understanding rather than merely memorizing the data. However, improvements plateau beyond certain ratios (e.g., 10× pairwise does not outperform 5×), and excessive imagewise augmentation can introduce distribution shifts that harm performance.

Hard masks outperformed soft masks, likely because hard masks fully replace masked regions with trainable parameters, while soft masks retain original image embeddings that can introduce noise and hinder learning. During inference, minimal or no masking yields the best performance, as richer input information enables more accurate predictions.

Our method outperformed most state-of-the-art models, confirming its generalization ability. A limitation is its reliance on image concatenation, which constrains input resolution and may reduce accuracy for fine structures. Future work will explore high-resolution segmentation while maintaining visual ICL benefits.

\section{Conclusions}
We proposed FlexICL, a novel visual ICL model for musculoskeletal US image segmentation. Although the SimMIM baseline model was less effective than others for a given data set size, by integrating multiple data augmentation strategies based on image concatenation, namely pairwise, imagewise, and epochwise augmentations, we significantly enhanced the diversity and volume of training data, which led to notable improvements in model performance to exceed even other fully-supervised models.

FlexICL outperformed most state-of-the-art models across various upper limb US datasets, demonstrating strong generalization. Even with a limited number of annotated frames, it can accurately generate annotations for the rest, reducing the annotation burden and enabling a scalable solution for clinical applications. This approach supports more accurate diagnosis and follow-up of musculoskeletal conditions, promoting wider adoption of US for fracture detection.

\section{Appendix}
\subsection{Data Statement}
The data used in this study contain sensitive patient information and are subject to privacy protection and ethics regulations approved by the institutional ethics board. As such, the data are available for internal research use only and cannot be shared publicly.
\subsection{Ethics approval}
The study was approved by University of Alberta Health Research Ethics Board—Biomedical Panel (Pro00077093) and all methods were performed in accordance with the relevant guidelines and regulations.
\subsection{Consent to participate}
Informed consent was obtained from all participants and their guardians for this study.
\subsection{Consent to publish}
The authors affirm that human research participants provided informed consent for the publication of the images in Figures 1-6.

\section*{Acknowledgment}
Dr. Jacob L. Jaremko is supported by the Canada CIFAR AI Chair, and his academic time is made available by Medical Imaging Consultants (MIC), Edmonton, Canada. We acknowledge the funding support of Alberta Innovates AICE Concepts (US-AID), AMII RAP, CIHR, Alberta Innovates Graduate Student Scholarships, IC-IMPACTS, WCHRI Graduate Studentship, TD Bank, AMII, CIFAR, and the Alberta Emergency Strategic Clinical Network. We thank the Digital Research Alliance of Canada for providing us with computational resources including high-power Graphical Processing Units (GPU) that were used for training and testing our deep learning models.

\bibliographystyle{IEEEtran}  
\bibliography{Zhou_reference}

\begin{thebibliography}{10}
\providecommand{\url}[1]{#1}
\csname url@samestyle\endcsname
\providecommand{\newblock}{\relax}
\providecommand{\bibinfo}[2]{#2}
\providecommand{\BIBentrySTDinterwordspacing}{\spaceskip=0pt\relax}
\providecommand{\BIBentryALTinterwordstretchfactor}{4}
\providecommand{\BIBentryALTinterwordspacing}{\spaceskip=\fontdimen2\font plus
\BIBentryALTinterwordstretchfactor\fontdimen3\font minus \fontdimen4\font\relax}
\providecommand{\BIBforeignlanguage}[2]{{%
\expandafter\ifx\csname l@#1\endcsname\relax
\typeout{** WARNING: IEEEtran.bst: No hyphenation pattern has been}%
\typeout{** loaded for the language `#1'. Using the pattern for}%
\typeout{** the default language instead.}%
\else
\language=\csname l@#1\endcsname
\fi
#2}}
\providecommand{\BIBdecl}{\relax}
\BIBdecl

\bibitem{xiemichaelAugmentingLowCostPoint2025}
{Xie, Michael}, {Zhou, Yuyue}, V.~Man, S.~McDonald, C.~Gallant, J.~Kupper, A.~Hareendranathan, and J.~L. Jaremko, ``Augmenting {{Low-Cost Point}} of {{Care Ultrasound}} with {{Interpretable Artificial Intelligence}} for {{Diagnosis}} of {{Pediatric Wrist Fracture}}.'' \emph{Canada {{Association}} of {{Radiologists}} 2025 {{Annual Scientific Meeting}} ({{CAR}} 2025)}, 2025.

\bibitem{knight2D3DUltrasound2023d}
J.~Knight, Y.~Zhou, C.~Keen, A.~R. Hareendranathan, F.~Alves-Pereira, S.~Ghasseminia, S.~Wichuk, A.~Brilz, D.~Kirschner, and J.~Jaremko, ``{{2D}}/{{3D}} ultrasound diagnosis of pediatric distal radius fractures by human readers vs artificial intelligence,'' \emph{Scientific Reports}, vol.~13, no.~1, p. 14535, 2023.

\bibitem{knight2DVs3D2023a}
J.~Knight, F.~Alves-Pereira, C.~E. Keen, and J.~L. Jaremko, ``{{2D}} vs. {{3D Ultrasound Diagnosis}} of {{Pediatric Supracondylar Fractures}},'' \emph{Children}, vol.~10, no.~11, p. 1766, 2023.

\bibitem{brudvikChildhoodFracturesBergen2003b}
C.~Brudvik and L.~M. Hove, ``Childhood fractures in {{Bergen}}, {{Norway}}: Identifying high-risk groups and activities,'' \emph{Journal of Pediatric Orthopedics}, vol.~23, no.~5, pp. 629--634, 2003.

\bibitem{hedstromEpidemiologyFracturesChildren2010c}
E.~M. Hedström, O.~Svensson, U.~Bergström, and P.~Michno, ``Epidemiology of fractures in children and adolescents,'' \emph{Acta Orthopaedica}, vol.~81, no.~1, pp. 148--153, 2010.

\bibitem{chengLimbFracturePattern1993b}
J.~C. Cheng and W.~Y. Shen, ``Limb fracture pattern in different pediatric age groups: A study of 3,350 children,'' \emph{Journal of Orthopaedic Trauma}, vol.~7, no.~1, pp. 15--22, 1993.

\bibitem{saeedElbowFracturesOverview2025}
W.~Saeed and M.~Waseem, ``Elbow {{Fractures Overview}},'' in \emph{{{StatPearls}}}.\hskip 1em plus 0.5em minus 0.4em\relax StatPearls Publishing, 2025.

\bibitem{skaggsElbowFracturesChildren1997a}
D.~L. Skaggs, ``Elbow {{Fractures}} in {{Children}}: {{Diagnosis}} and {{Management}},'' \emph{The Journal of the American Academy of Orthopaedic Surgeons}, vol.~5, no.~6, pp. 303--312, 1997.

\bibitem{pinto-coelhoHowArtificialIntelligence2023}
L.~Pinto-Coelho, ``How {{Artificial Intelligence Is Shaping Medical Imaging Technology}}: {{A Survey}} of {{Innovations}} and {{Applications}},'' \emph{Bioengineering}, vol.~10, no.~12, p. 1435, 2023.

\bibitem{kirillovSegmentAnything2023a}
A.~Kirillov, E.~Mintun, N.~Ravi, H.~Mao, C.~Rolland, L.~Gustafson, T.~Xiao, S.~Whitehead, A.~C. Berg, W.-Y. Lo, P.~Dollár, and R.~Girshick, ``Segment {{Anything}},'' in \emph{2023 {{IEEE}}/{{CVF International Conference}} on {{Computer Vision}} ({{ICCV}})}.\hskip 1em plus 0.5em minus 0.4em\relax IEEE, 2023, pp. 3992--4003.

\bibitem{tajbakhshEmbracingImperfectDatasets2020}
N.~Tajbakhsh, L.~Jeyaseelan, Q.~Li, J.~N. Chiang, Z.~Wu, and X.~Ding, ``Embracing imperfect datasets: {{A}} review of deep learning solutions for medical image segmentation,'' \emph{Medical Image Analysis}, vol.~63, p. 101693, 2020.

\bibitem{guanDomainAdaptationMedical2022}
H.~Guan and M.~Liu, ``Domain {{Adaptation}} for {{Medical Image Analysis}}: {{A Survey}},'' \emph{IEEE transactions on bio-medical engineering}, vol.~69, no.~3, pp. 1173--1185, 2022.

\bibitem{alzubaidiNovelTransferLearning2021}
L.~Alzubaidi, M.~Al-Amidie, A.~Al-Asadi, A.~J. Humaidi, O.~Al-Shamma, M.~A. Fadhel, J.~Zhang, J.~Santamaría, and Y.~Duan, ``Novel {{Transfer Learning Approach}} for {{Medical Imaging}} with {{Limited Labeled Data}},'' \emph{Cancers}, vol.~13, no.~7, p. 1590, 2021.

\bibitem{zoetmulderDomainTaskspecificTransfer2022}
R.~Zoetmulder, E.~Gavves, M.~Caan, and H.~Marquering, ``Domain- and task-specific transfer learning for medical segmentation tasks,'' \emph{Computer Methods and Programs in Biomedicine}, vol. 214, p. 106539, 2022.

\bibitem{albadawyDeepLearningSegmentation2018}
E.~A. AlBadawy, A.~Saha, and M.~A. Mazurowski, ``Deep learning for segmentation of brain tumors: {{Impact}} of cross-institutional training and testing,'' \emph{Medical Physics}, vol.~45, no.~3, pp. 1150--1158, 2018.

\bibitem{poochCanWeTrust2020}
E.~H.~P. Pooch, P.~L. Ballester, and R.~C. Barros, ``Can we trust deep learning models diagnosis? the impact of domain shift in chest radiograph classification,'' \emph{arXiv preprint arXiv:1909.01940}, 2020.

\bibitem{kimDistributionAligningRefinery2021}
J.~Kim, Y.~Hur, S.~Park, E.~Yang, S.~J. Hwang, and J.~Shin, ``Distribution aligning refinery of pseudo-label for imbalanced semi-supervised learning,'' \emph{arXiv preprint arXiv:2007.08844}, 2021.

\bibitem{wuExploringFeatureRepresentation2024}
H.~Wu, X.~Li, and K.-T. Cheng, ``Exploring {{Feature Representation Learning}} for {{Semi-supervised Medical Image Segmentation}},'' \emph{IEEE Transactions on Neural Networks and Learning Systems}, vol.~35, no.~11, pp. 16\,589--16\,601, 2024.

\bibitem{liuSemiSupervisedMedicalImage2024}
K.~Liu, S.~Ling, and S.~Liu, ``Semi-{{Supervised Medical Image Classification}} with {{Pseudo Labels Using Coalition Similarity Training}},'' \emph{Mathematics}, vol.~12, no.~10, p. 1537, 2024.

\bibitem{zengSSTBNSemiSupervisedTriBranch2023}
L.-L. Zeng, K.~Gao, D.~Hu, Z.~Feng, C.~Hou, P.~Rong, and W.~Wang, ``{{SS-TBN}}: {{A Semi-Supervised Tri-Branch Network}} for {{COVID-19 Screening}} and {{Lesion Segmentation}},'' \emph{IEEE Transactions on Pattern Analysis and Machine Intelligence}, vol.~45, no.~8, pp. 10\,427--10\,442, 2023.

\bibitem{yaoEnhancingPseudoLabel2022}
H.~Yao, X.~Hu, and X.~Li, ``Enhancing pseudo label quality for semi-supervised domain-generalized medical image segmentation,'' \emph{arXiv preprint arXiv:2201.08657}, 2022.

\bibitem{nieCrossDomainFewShotSegmentation2024}
J.~Nie, Y.~Xing, G.~Zhang, P.~Yan, A.~Xiao, Y.-P. Tan, A.~C. Kot, and S.~Lu, ``Cross-{{Domain Few-Shot Segmentation}} via {{Iterative Support-Query Correspondence Mining}},'' 2024, pp. 3380--3390.

\bibitem{balestrieroCookbookSelfSupervisedLearning2023}
R.~Balestriero, M.~Ibrahim, V.~Sobal, A.~Morcos, S.~Shekhar, T.~Goldstein, F.~Bordes, A.~Bardes, G.~Mialon, Y.~Tian, A.~Schwarzschild, A.~G. Wilson, J.~Geiping, Q.~Garrido, P.~Fernandez, A.~Bar, H.~Pirsiavash, Y.~LeCun, and M.~Goldblum, ``A cookbook of self-supervised learning,'' \emph{arXiv preprint arXiv:2304.12210}, 2023.

\bibitem{felfeliyanSelfsupervisedRCNNMedicalImage2023b}
B.~Felfeliyan, N.~D. Forkert, A.~Hareendranathan, D.~Cornel, Y.~Zhou, G.~Kuntze, J.~L. Jaremko, and J.~L. Ronsky, ``Self-supervised-{{RCNN}} for medical image segmentation with limited data annotation,'' \emph{Computerized Medical Imaging and Graphics}, vol. 109, p. 102297, 2023.

\bibitem{almalkiSelfSupervisedLearningMasked2023a}
A.~Almalki and L.~J. Latecki, ``Self-{{Supervised Learning}} with {{Masked Image Modeling}} for {{Teeth Numbering}}, {{Detection}} of {{Dental Restorations}}, and {{Instance Segmentation}} in {{Dental Panoramic Radiographs}},'' in \emph{2023 {{IEEE}}/{{CVF Winter Conference}} on {{Applications}} of {{Computer Vision}} ({{WACV}})}.\hskip 1em plus 0.5em minus 0.4em\relax IEEE, 2023, pp. 5583--5592.

\bibitem{zaiemPretextTasksSelection2021}
S.~Zaiem, T.~Parcollet, S.~Essid, and A.~Heba, ``Pretext {{Tasks Selection}} for {{Multitask Self-Supervised Speech Representation Learning}},'' 2021.

\bibitem{albelwiSurveySelfSupervisedLearning2022}
S.~Albelwi, ``Survey on {{Self-Supervised Learning}}: {{Auxiliary Pretext Tasks}} and {{Contrastive Learning Methods}} in {{Imaging}},'' \emph{Entropy}, vol.~24, no.~4, p. 551, 2022.

\bibitem{zhouSelfSupervisedLearningMore2023c}
Y.~Zhou, J.~Knight, B.~Felfeliyan, S.~Ghosh, F.~Alves-Pereira, C.~Keen, A.~R. Hareendranathan, and J.~L. Jaremko, ``Self-{{Supervised Learning}} to~{{More Efficiently Generate Segmentation Masks}} for~{{Wrist Ultrasound}},'' in \emph{Simplifying {{Medical Ultrasound}}}, B.~Kainz, A.~Noble, J.~Schnabel, B.~Khanal, J.~P. Müller, and T.~Day, Eds.\hskip 1em plus 0.5em minus 0.4em\relax Springer Nature Switzerland, pp. 79--88.

\bibitem{raviSAMSegmentAnything2024}
N.~Ravi, V.~Gabeur, Y.-T. Hu, R.~Hu, C.~Ryali, T.~Ma, H.~Khedr, R.~Rädle, C.~Rolland, L.~Gustafson, E.~Mintun, J.~Pan, K.~V. Alwala, N.~Carion, C.-Y. Wu, R.~Girshick, P.~Doll{\'a}r, and C.~Feichtenhofer, ``Sam 2: Segment anything in images and videos,'' \emph{arXiv preprint arXiv:2408.00714}, 2024.

\bibitem{heComputerVisionBenchmarkSegmentAnything2023a}
S.~He, R.~Bao, J.~Li, J.~Stout, A.~Bjornerud, P.~E. Grant, and Y.~Ou, ``Computer-vision benchmark segment-anything model (sam) in medical images: Accuracy in 12 datasets,'' \emph{arXiv preprint arXiv:2304.09324}, 2023.

\bibitem{huangSegmentAnythingModel2024}
Y.~Huang, X.~Yang, L.~Liu, H.~Zhou, A.~Chang, X.~Zhou, R.~Chen, J.~Yu, J.~Chen, C.~Chen, S.~Liu, H.~Chi, X.~Hu, K.~Yue, L.~Li, V.~Grau, D.-P. Fan, F.~Dong, and D.~Ni, ``Segment anything model for medical images?'' \emph{Medical Image Analysis}, vol.~92, p. 103061, 2024.

\bibitem{ronnebergerUNetConvolutionalNetworks2015d}
O.~Ronneberger, P.~Fischer, and T.~Brox, ``U-{{Net}}: {{Convolutional Networks}} for {{Biomedical Image Segmentation}},'' in \emph{Medical {{Image Computing}} and {{Computer-Assisted Intervention}} – {{MICCAI}} 2015}, ser. Lecture {{Notes}} in {{Computer Science}}, N.~Navab, J.~Hornegger, W.~M. Wells, and A.~F. Frangi, Eds.\hskip 1em plus 0.5em minus 0.4em\relax Springer International Publishing, 2015, pp. 234--241.

\bibitem{bardisDeepLearningLimited2020}
M.~Bardis, R.~Houshyar, C.~Chantaduly, A.~Ushinsky, J.~Glavis-Bloom, M.~Shaver, D.~Chow, E.~Uchio, and P.~Chang, ``Deep {{Learning}} with {{Limited Data}}: {{Organ Segmentation Performance}} by {{U-Net}},'' \emph{Electronics}, vol.~9, no.~8, p. 1199, 2020.

\bibitem{zhangInstructMeMore2024b}
J.~Zhang, B.~Wang, L.~Li, Y.~Nakashima, and H.~Nagahara, ``Instruct {{Me More}}! {{Random Prompting}} for {{Visual In-Context Learning}},'' 2024, pp. 2597--2606.

\bibitem{bahngExploringVisualPrompts2022b}
H.~Bahng, A.~Jahanian, S.~Sankaranarayanan, and P.~Isola, ``Exploring visual prompts for adapting large-scale models,'' \emph{arXiv preprint arXiv:2203.17274}, 2022.

\bibitem{huangPromptingUnseenDetecting2024}
Z.-X. Huang, J.-W. Chen, Z.-P. Zhang, and C.-M. Yu, ``Prompting the unseen: Detecting hidden backdoors in black-box models,'' \emph{arXiv preprint arXiv:2411.09540}, 2024.

\bibitem{renAreYouCopying2024}
H.~Ren, A.~Yan, C.~z.~Gao, H.~Yan, Z.~Zhang, and J.~Li, ``Are you copying my prompt? protecting the copyright of vision prompt for vpaas via watermark,'' \emph{arXiv preprint arXiv:2405.15161}, 2024.

\bibitem{wuUnleashingPowerVisual2023a}
J.~Wu, X.~Li, C.~Wei, H.~Wang, A.~Yuille, Y.~Zhou, and C.~Xie, ``Unleashing the power of visual prompting at the pixel level,'' \emph{arXiv preprint arXiv:2212.10556}, 2023.

\bibitem{hossainVisualPromptingGeneralized2024}
M.~R.~I. Hossain, M.~Siam, L.~Sigal, and J.~J. Little, ``Visual {{Prompting}} for {{Generalized Few-shot Segmentation}}: {{A Multi-scale Approach}},'' 2024, pp. 23\,470--23\,480.

\bibitem{zhaoInstructBrushLearningAttentionbased2024}
R.~Zhao, Q.~Fan, F.~Kou, S.~Qin, H.~Gu, W.~Wu, P.~Xu, M.~Zhu, N.~Wang, and X.~Gao, ``Instructbrush: Learning attention-based instruction optimization for image editing,'' \emph{arXiv preprint arXiv:2403.18660}, 2024.

\bibitem{vettoruzzoUnsupervisedMetaLearningContext2025}
A.~Vettoruzzo, L.~Braccaioli, J.~Vanschoren, and M.~Nowaczyk, ``Unsupervised meta-learning via in-context learning,'' \emph{arXiv preprint arXiv:2405.16124}, 2025.

\bibitem{rajagopalanAWRaCLeAllWeatherImage2024}
S.~Rajagopalan and V.~M. Patel, ``Awracle: All-weather image restoration using visual in-context learning,'' \emph{arXiv preprint arXiv:2409.00263}, 2024.

\bibitem{barVisualPromptingImage}
A.~Bar, Y.~Gandelsman, T.~Darrell, A.~Globerson, and A.~A. Efros, ``Visual {{Prompting}} via {{Image Inpainting}}.''

\bibitem{wangImagesSpeakImages2023b}
X.~Wang, W.~Wang, Y.~Cao, C.~Shen, and T.~Huang, ``Images {{Speak}} in {{Images}}: {{A Generalist Painter}} for {{In-Context Visual Learning}},'' in \emph{Proceedings of the {{IEEE}}/{{CVF Conference}} on {{Computer Vision}} and {{Pattern Recognition}}}, 2023, pp. 6830--6839.

\bibitem{dosovitskiyImageWorth16x162021c}
A.~Dosovitskiy, L.~Beyer, A.~Kolesnikov, D.~Weissenborn, X.~Zhai, T.~Unterthiner, M.~Dehghani, M.~Minderer, G.~Heigold, S.~Gelly, J.~Uszkoreit, and N.~Houlsby, ``An image is worth 16x16 words: Transformers for image recognition at scale,'' \emph{arXiv preprint arXiv:2010.11929}, 2021.

\bibitem{sunExploringEffectiveFactors2023b}
Y.~Sun, Q.~Chen, J.~Wang, J.~Wang, and Z.~Li, ``Exploring effective factors for improving visual in-context learning,'' \emph{arXiv preprint arXiv:2304.04748}, 2023.

\bibitem{wangSegGPTSegmentingEverything2023b}
X.~Wang, X.~Zhang, Y.~Cao, W.~Wang, C.~Shen, and T.~Huang, ``{{SegGPT}}: {{Towards Segmenting Everything}} in {{Context}},'' 2023, pp. 1130--1140.

\bibitem{liuUnifyingImageProcessing2024a}
Y.~Liu, X.~Chen, X.~Ma, X.~Wang, J.~Zhou, Y.~Qiao, and C.~Dong, ``Unifying image processing as visual prompting question answering,'' \emph{arXiv preprint arXiv:2310.10513}, 2024.

\bibitem{zhangWhatMakesGood2023b}
Y.~Zhang, K.~Zhou, and Z.~Liu, ``What {{Makes Good Examples}} for {{Visual In-Context Learning}}?'' \emph{Advances in Neural Information Processing Systems}, vol.~36, pp. 17\,773--17\,794, 2023.

\bibitem{xuGlobalOptimalVisual2024}
C.~Xu, C.~Liu, Y.~Wang, Y.~Yao, and Y.~Fu, ``Towards global optimal visual in-context learning prompt selection,'' \emph{arXiv preprint arXiv:2405.15279}, 2024.

\bibitem{kumarVisualContextLearning2024}
N.~Kumar, O.~Aran, and V.~Vasudevan, ``Visual {{In-Context Learning}} for {{Few-Shot Eczema Segmentation}},'' in \emph{2024 46th {{Annual International Conference}} of the {{IEEE Engineering}} in {{Medicine}} and {{Biology Society}} ({{EMBC}})}, 2024, pp. 1--5.

\bibitem{wuTumorSegmentationWhole2024}
H.~Wu, C.~Br{\'e}mond-Martin, K.~Bouaou, and C.~Clouchoux, ``Tumor segmentation on whole slide images: Training or prompting?'' \emph{arXiv preprint arXiv:2402.13932}, 2024.

\bibitem{zhouSimICLSimpleVisual2024}
Y.~Zhou, B.~Felfeliyan, S.~Ghosh, J.~Knight, F.~Alves-Pereira, C.~Keen, J.~Küpper, A.~R. Hareendranathan, and J.~L. Jaremko, ``{{SimICL}}: {{A Simple Visual In-context Learning Framework}} for {{Ultrasound Segmentation}},'' in \emph{2024 46th {{Annual International Conference}} of the {{IEEE Engineering}} in {{Medicine}} and {{Biology Society}} ({{EMBC}})}, 2024, pp. 1--4.

\bibitem{xieSimMIMSimpleFramework2022b}
Z.~Xie, Z.~Zhang, Y.~Cao, Y.~Lin, J.~Bao, Z.~Yao, Q.~Dai, and H.~Hu, ``{{SimMIM}}: {{A Simple Framework}} for {{Masked Image Modeling}},'' 2022, pp. 9653--9663.

\bibitem{chenTransUNetTransformersMake2021}
J.~Chen, Y.~Lu, Q.~Yu, X.~Luo, E.~Adeli, Y.~Wang, L.~Lu, A.~L. Yuille, and Y.~Zhou, ``Transunet: Transformers make strong encoders for medical image segmentation,'' \emph{arXiv preprint arXiv:2102.04306}, 2021.

\end{thebibliography}

\end{document}